\documentclass[10pt,journal,compsoc]{IEEEtran}
\ifCLASSOPTIONcompsoc
  \usepackage[nocompress]{cite}
\else
  \usepackage{cite}
\fi
\ifCLASSINFOpdf
\else
\fi
\usepackage{url}
\usepackage{graphicx}
\usepackage{booktabs}
\usepackage[normalem]{ulem}
\usepackage{fvextra}
\usepackage{fancyvrb}
\usepackage{ragged2e}

\usepackage{amsmath}
\usepackage{mathtools}
\usepackage{amsthm}

\usepackage{pifont}       % \ding{xx}
\usepackage{bbding}       % \Checkmark,\XSolid,... (需要和pifont宏包共同使用)
\usepackage{fontawesome}
\usepackage{xspace}

\usepackage{float}

\usepackage{algorithmic}
\usepackage{algorithm}
\usepackage{amssymb}

\usepackage[table]{xcolor}     % row colors (load before tcolorbox to avoid option clash)
\usepackage[most]{tcolorbox}
\usepackage{caption}
\usepackage{siunitx}           % S column for number alignment
\usepackage{makecell}          % optional, for multi-line cell if needed
\begin{document}
\title{Sci-CoE: Co-evolving Scientific Reasoning LLMs via Geometric Consensus with Sparse Supervision}

\author{Xiaohan~He, Shiyang~Feng, Songtao~Huang, Zongsheng~Cao, Bin~Wang, and Bo~Zhang% <-this % stops a space
\IEEEcompsocitemizethanks{\IEEEcompsocthanksitem Xiaohan He and Bin Wang are with School of Future Information Technology, Fudan University.
\IEEEcompsocthanksitem Shiyang Feng, Songtao Huang, Zongsheng Cao, and Bo Zhang are with Shanghai Artificial Intelligence Laboratory, Shanghai 200232, China. (Corresponding author: Shiyang Feng and Bo Zhang, E-mail: fengshiyang@pjlab.org.cn and zhangbo@pjlab.org.cn).
}}% <-this % stops a space
\IEEEtitleabstractindextext{
\begin{abstract}
\justifying{
Large language models (LLMs) have demonstrated exceptional reasoning capabilities, and co-evolving paradigms have shown promising results in domains with verifiable reward signals, such as code and math. However, in unverifiable domains, such as scientific reasoning, these methods remain fragile due to unreliable solution evaluation and limited diversity in verification strategies. In this work, we propose Sci-CoE, a two-stage co-evolving framework for domains without verifiable feedback. The framework enables models to self-evolve as both solvers and verifiers by transitioning from sparse supervision to unsupervised learning. In the first stage, the model uses a small set of annotated data to establish fundamental correctness judgment anchors for the Verifier. In the second stage, we introduce a geometric reward mechanism that jointly considers consensus, reliability, and diversity, driving large-scale self-iteration on unlabeled data. Experiments on several general scientific benchmarks demonstrate that Sci-CoE enhances complex reasoning capabilities and exhibits strong scalability, facilitating the construction of more robust and diverse evaluation systems. Codes are available at \url{https://github.com/InternScience/Sci-CoE}.
}
\end{abstract}

\begin{IEEEkeywords}
Scientific Reasoning, Large Language Models, Co-evolution, Few Supervision
\end{IEEEkeywords}
}
\maketitle

% is not selected <OR> if conference mode is selected - because compsoc
% conference papers position the abstract like regular (non-compsoc)
\IEEEdisplaynontitleabstractindextext

% \fi
\IEEEpeerreviewmaketitle

\section{Introduction}
\label{sec:intro}

\IEEEPARstart{S}{elf-evolving} Reinforcement Learning (RL) has emerged as a transformative paradigm for Large Language Models (LLMs) to refine their reasoning trajectories through feedback. A notable milestone is the Zero RL paradigm introduced by DeepSeek-R1~\cite{guo2025deepseekr1}, which elicits sophisticated reasoning behaviors without prior supervised imitation learning. However, this approach remains fundamentally dependent on annotated datasets for reward calculation. To mitigate this reliance, self-play mechanisms have been adopted to facilitate autonomous evolution. In these settings, the LLM concurrently assumes multiple roles, such as a challenger and a solver~\cite{huang2025r-zero, zhao2025absolute-zero} or a solver and a verifier~\cite{wang2025cure}, to drive co-evolution through mutual interaction. These self-play methods significantly enhance the autonomy of reasoning training by reducing the need for external supervision. \par  

\begin{figure}[t]
    \centering
    \small
    \resizebox{1.0\linewidth}{!}{\includegraphics{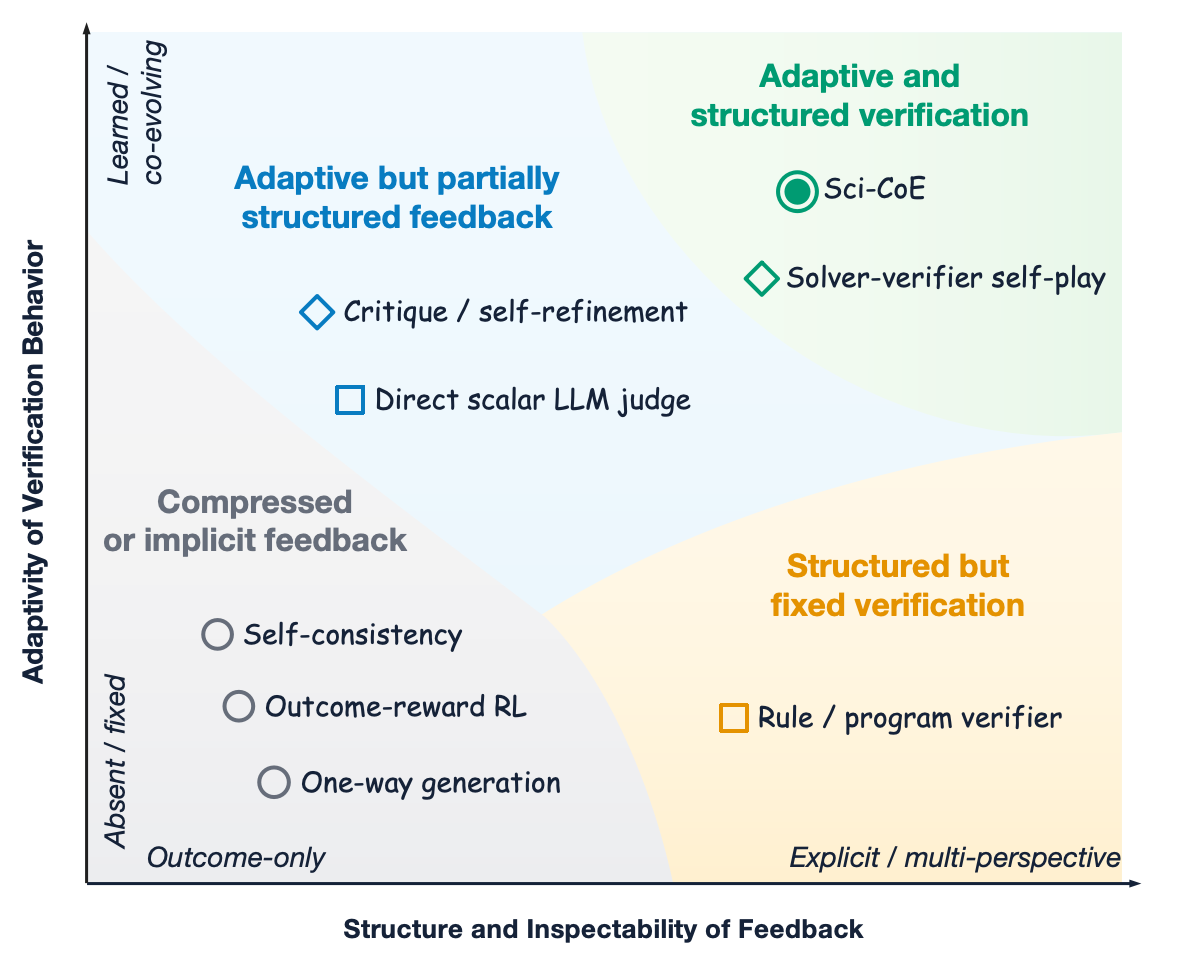}}
    \caption{Positioning of feedback paradigms for scientific reasoning along two axes: the structure and inspectability of the feedback (horizontal) and the adaptivity of the verification behavior (vertical). }
    \label{fig:positioning}
\vspace{-2pt}
\end{figure}

However, existing self‑evolution paradigms are primarily confined to domains such as coding and mathematics, where task quality can be assessed through clear verification signals. In these domains, correctness can either be judged directly using ground‑truth solutions or indirectly through explicit verification methods such as unit tests. In contrast, a broader class of unverifiable domains does not provide such task-level verifiable feedback, making autonomous self-evolution far more challenging. Scientific reasoning is an important and representative instance of this setting. Specifically, these tasks unfold in an open‑world regime with multiple valid solutin pathways and heterogeneous verification criteria. Furthermore, scientific reasoning tasks spans diverse disciplines where verification requires expert assessment of complex intermediate logic rather than simple answer matching. The prohibitive cost of curating such specialized supervision renders large-scale and data-intensive training approaches impractical. Motivated by these challenges, we use scientific reasoning as a representative and challenging experimental setting, and raise a pivotal question:

\textit{\textbf{Can we develop a self-evolving RL framework for domains without verifiable feedback under limited supervision?}} \par

A closer look reveals that the difficulty is not only a shortage of labels, but also the quality of the feedback loop itself (Figure~\ref{fig:positioning}). A one-way generator receives no process-level corrective signal. A model used as a direct scalar judge compresses a rich scientific evaluation into an opaque decision that carries judge-specific noise and bias. And when the policy also produces its own verification criteria, correlated errors, overly permissive checks, and repeated strategies can yield high agreement without genuine discrimination---a collapse that starves the solver of an informative reward. Effective self-evolution in domains without verifiable feedback therefore requires verification feedback that is at once adaptive and structured.\par

In this work, we introduce Sci‑CoE, a two-stage co-evolving framework for domains without verifiable feedback. Under this framework, a single model simultaneously acts as both a Solver and a Verifier. The Solver generates candidate solutions, and the Verifier constructs strategies to evaluate their correctness. These two roles are optimized jointly through interactive reinforcement learning. Furthermore, to maintain stability without ground‑truth supervision, we propose a geometric reward mechanism that represents verification strategies in a latent geometric space. This mechanism encourages strategies to remain both reliable and diverse, which prevents consensus collapse and supports sustained self‑evolution in open‑ended scientific domains.\par

\begin{figure}[t]
    \centering
    \small
    \resizebox{1.0\linewidth}{!}{\includegraphics{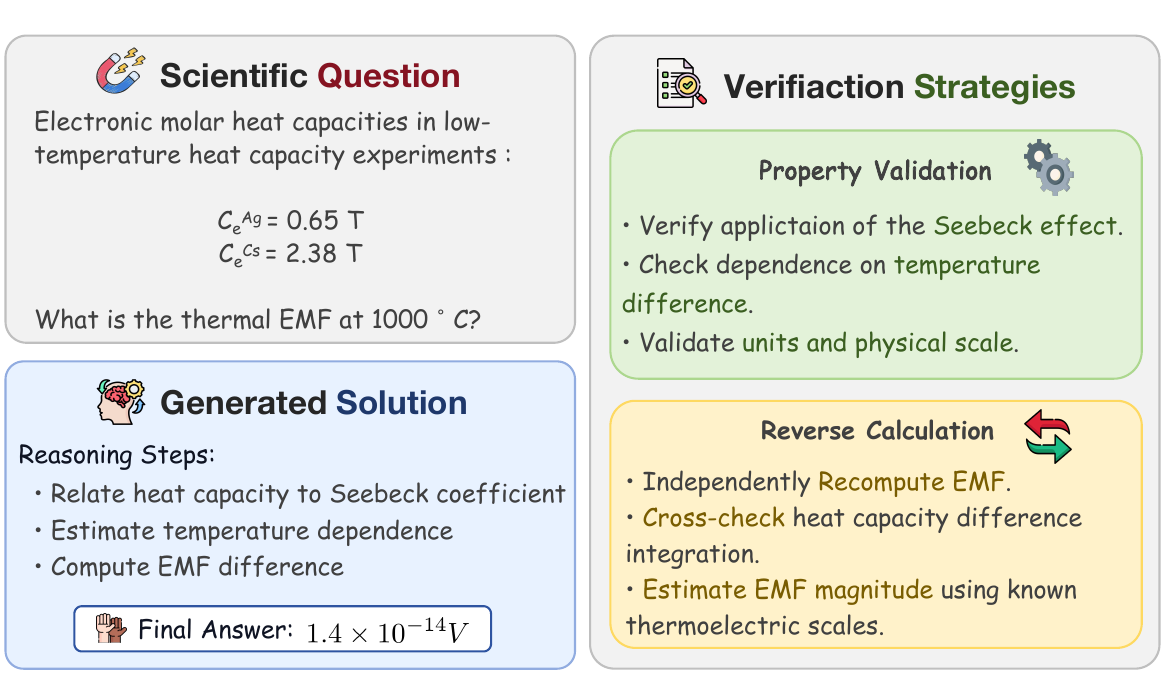}}
    \caption{An example scientific question with a generated candidate solution and multiple natural-language verification strategies. Each strategy examines a different aspect of the solution, such as validating physical properties or performing an independent reverse calculation, rather than directly producing another answer.}
    \label{fig:example}
\vspace{-2pt}
\end{figure}

Comprehensive experimental results demonstrate that Sci‑CoE remains effective even when explicit verification signals are absent. Across diverse scientific domains, the framework achieves strong reasoning performance under limited supervision. Furthermore, the proposed reward mechanism enhances both the reliability and the diversity of verification strategies, leading to consistent performance gains. Our contributions are summarized as follows:

1) We proposed a co‑evolving framework called Sci‑CoE for scientific reasoning that integrates a Solver and a Verifier within a single LLM. This design enables the acquisition of both solution‑generation and solution‑verification capabilities, supporting self‑evolution without ground‑truth solutions or predefined verification procedures.

2) A geometric reward mechanism is developed to model verification strategies in a latent geometric space. By encouraging both reliability and diversity, this mechanism prevents consensus collapse and enables stable unsupervised evolution.

3) Comprehensive experiments show that Sci‑CoE framework not only improves reasoning accuracy and robustness but also cultivates effective verification behaviors capable of multi‑perspective evaluation of scientific questions.

\section{Related Work}

\subsection{Scientific Reasoning with LLMs} 
The rapid progress of large language models (LLMs) has spurred a systematic transition from fast, pattern-matching inference toward slow, deliberate reasoning, a shift that recent surveys characterize as moving from ``System~1'' to ``System~2'' behavior~\cite{li2025system2survey}. Along this trajectory, scientific LLMs have been advanced through broad scientific pretraining, curated reasoning data, multimodal modeling, and domain-specific adaptation~\cite{hu2025survey,team2025novelseek,fallahpour2025bioreason,tan2025chemmllm,zhang2024chemllm}. Intern-S1 utilizes a multimodal Mixture-of-Experts (MoE) architecture with Mixture-of-Rewards reinforcement learning to outperform closed-source models in complex scientific tasks~\cite{bai2025intern-s1}. SciReasoner aligns natural language with heterogeneous scientific representations to establish a reasoning foundation~\cite{wang2025scireasoner}, while SCI-Verifier introduces a unified reasoning-augmented framework for robust equivalence judgment in verification~\cite{zheng2025sciverify}. Regarding domain-specific innovations, ChemVLM integrates visual encoders to bridge molecular structures with text~\cite{li2025chemvlm}. Med-R1 applies reinforcement learning for medical vision-language reasoning~\cite{lai2025medr1}, and MindLLM employs a subject-agnostic framework to decode fMRI signals directly into text~\cite{qiu2025mindllm}. AstroMLab 3 leverages high-quality data curation to enable a compact 8B-parameter model to match GPT-4o performance in astronomy~\cite{de2024astromlab}. Rather than improving a single capability in isolation, Sci-CoE focuses on the interaction between solving and verifying, training one policy to generate both scientific solutions and explicit verification strategies.\par

\begin{table*}[t!]
\centering
\small
\setlength{\tabcolsep}{3.5pt}
\caption{\textbf{Conceptual comparison of feedback paradigms for reasoning.} ``Task checker'' denotes a task-specific executable or exact verifier; ``Explicit strategies'' denotes an inspectable, natural-language verification procedure; ``Diversity control'' denotes an explicit mechanism against homogeneous verification.}
\label{tab:method_positioning}
\begin{tabular}{lccccc}
\toprule
\textbf{Paradigm} & \textbf{Sparse labels} & \textbf{Task checker} & \textbf{Explicit strategies} & \textbf{Diversity control} & \textbf{Unlabeled adaptation} \\
\midrule
RLHF / RLAIF & Required & No & No & No & Limited \\
RLVR & Not required & Required & No & No & Yes \\
Direct LLM-as-a-Judge & No & No & No & No & Yes \\
Solver--Verifier self-play & Varies & Usually & Varies & Limited & Varies \\
\rowcolor{gray!12}\textbf{Sci-CoE (ours)} & Stage~1 only & No & Yes & Geometric & Yes (Stage~2) \\
\bottomrule
\end{tabular}
\end{table*}

\subsection{Self-Evolving Large Language Models}
Early research in self-evolution LLMs explored the self-play between generation and verification, particularly within code domains. Sol-Ver introduces a self-play framework where models iteratively refine both code implementations and test cases~\cite{lin2025learning}. CURE leverages reinforcement learning to mutually enhance the LLM coder and unit tester through dynamic interaction~\cite{wang2025cure}. Pushing autonomy further, subsequent frameworks learn to generate their own problems and adaptive curricula from scratch or minimal seeds.  Absolute Zero demonstrates that reasoning capabilities can emerge purely through reinforced self-play without human priors~\cite{zhao2025absolute-zero}, whereas SERL bootstraps robust policies from limited data via iterative selection~\cite{fang2025serl}. Scaling this verification paradigm, Loong introduces an agent-environment loop that synthesizes large-scale training data with executable code-based ground truth, enabling Reinforcement Learning with Verifiable Rewards (RLVR) across diverse disciplines~\cite{huang2025loong}. R-Zero employs internal consistency as a reward signal to facilitate self-evolution in general reasoning domains where external verifiers are absent \cite{huang2025r-zero}. Addressing the instability of such open-ended exploration, R-Few introduces a guided self-play mechanism to explicitly mitigate concept drift and diversity collapse \cite{yu2025r-few}. Distinct from these approaches, our work explores self-evolution for general scientific reasoning with limited data. We orchestrate the co-evolution of reasoning and rigorous verification strategies to eliminate dependencies on external verifiers while establishing a closed-loop reinforcement of scientific logic.\par

\subsection{Learning from Feedback and Verifiable Rewards}
Reinforcement learning from human or AI feedback typically converts preferences or model judgments into scalar rewards for policy optimization~\cite{ouyang2022training,bai2022constitutional}. Reinforcement learning with verifiable rewards (RLVR) instead exploits exact answers, unit tests, or other deterministic signals, and has been highly effective for mathematical and code reasoning~\cite{shao2024deepseekmath,guo2025deepseekr1}. These paradigms become far less direct for scientific questions, whose intermediate logic and final responses require semantic evaluation. Process-oriented verification offers finer-grained, step-level feedback~\cite{lightman2023lets}, and the LLM-as-a-Judge paradigm offers scalable but model-dependent evaluation~\cite{zheng2023judging}. Sci-CoE does not discard model-based evaluation; instead it \emph{structures} it, first generating an explicit verification strategy, then executing that strategy on candidate solutions with a fixed judging model, and finally aggregating the resulting evaluation matrix through consensus and geometric rewards. This design keeps the scalability of model-based feedback while making the verification behavior explicit, inspectable, and regularizable.\par

\subsection{Comparative Analysis and Positioning}
Table~\ref{tab:method_positioning} contrasts Sci-CoE with representative feedback paradigms. RLHF/RLAIF and direct LLM-as-a-Judge learn from scalar preferences or judgments without explicitly modeling the verification procedure. RLVR provides reliable rewards but presupposes a task-specific checker. Existing solver--verifier self-play improves autonomy, yet it usually relies on executable environments or on unregularized internal agreement that is prone to diversity collapse. Sci-CoE occupies a complementary setting: sparse labels anchor the two roles in Stage~1, while Stage~2 combines unlabeled scientific questions, explicit natural-language strategies, strategy-conditioned execution, and geometry-aware regularization. Its aim is therefore not to eliminate external computation, but to obtain a more structured and less collapse-prone learning signal precisely when deterministic scientific verifiers are unavailable.\par

\section{Methodology}

\begin{figure*}[t]
    \centering
    \includegraphics[width=0.98\linewidth]{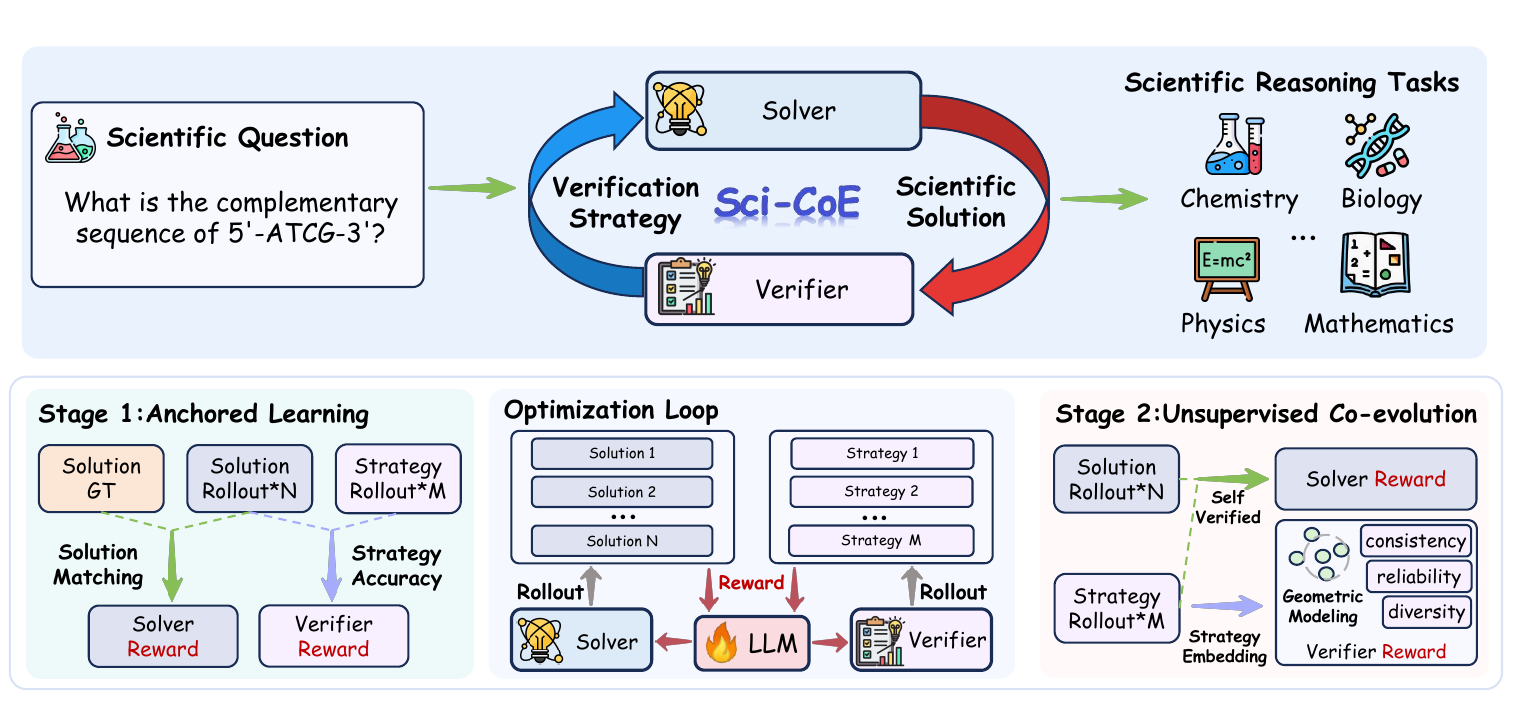}
    \vspace{-4pt}
    \caption{\textbf{Overall pipeline of Sci-CoE.} A single shared policy is prompted as Solver to generate candidate solutions and as Verifier to generate natural-language verification strategies. Stage 1 anchors both roles with a small labeled set; Stage 2 co-evolves them on unlabeled questions using consensus and geometric rewards.}
    \label{fig:framework}
\end{figure*}

\subsection{Overview}
We propose Sci-CoE, a Scientific Co-Evolving Framework designed to systematically improve scientific reasoning capabilities under minimal supervision, featuring a Solver and a Verifier. Training proceeds in two stages. In the first anchored learning stage, Sci-CoE leverages a small amount of labeled data to establish anchored notions of correctness and verification reliability, providing a stable initialization for both roles (Section \ref{sec: stage1}). In the second unsupervised co-evolution stage, the framework scales to large unlabeled data, where Solver and Verifier mutually supervise each other through a consensus and geometric reward mechanism, enabling co-evolution without task-level answer labels (Section \ref{sec: stage2}).\par
\subsection{The Solver-Verifier Co-Evolving Mechanism}
The key challenge we address is that, in scientific domains, solution generation and answer verification are both difficult and mutually dependent, especially when reliable ground-truth annotations are scarce. Sci-CoE is built upon a central insight: robust scientific reasoning emerges from the co-evolution of solution generation and verification capability. Instead of treating solving and evaluation as separate components, Sci-CoE trains a single model to simultaneously assume two complementary roles:
(1) \textbf{Solver}, which generates candidate solutions for scientific questions;
(2) \textbf{Verifier}, which generates verification strategies to assess the correctness of the solutions.\par

As illustrated in Figure \ref{fig:framework}, given a scientific question $q$, the policy model concurrently generates multiple solutions and multiple verification strategies:
\begin{equation}
S(q) = \{ s_1, \ldots, s_N \}, \quad V(q) = \{ v_1, \ldots, v_M \}
\end{equation}
The Solver produces solutions $s_i$ that contain explicit reasoning steps and a final answer, while the Verifier generates natural-language verification strategies $v_j$ that evaluate solution correctness from diverse perspectives, such as logical consistency checks, physical constraints, or inverse derivations.
Each solution–verification pair $(s_i,v_j)$ is evaluated by a fixed external LLM acting as a judging model, which strictly follows the specified verification strategy to assess the solution and outputs a binary result:
\begin{equation}
\text{Eval}(s_i, v_j) \in \{0, 1\}
\end{equation}
These results form a verification matrix $E\in \{ 0, 1 \}^{N \times M}$, which serves as the core feedback signal for reinforcement learning. The Solver and Verifier share the same set of model parameters and are jointly optimized using Proximal Policy Optimization (PPO).\par

This framework establishes a closed-loop co-evolving process: higher-quality solutions facilitate the learning of more discriminative verification strategies, while stronger verification strategies provide more reliable reward signals for solution generation in turn.\par

\subsection{Anchored Learning with Sparse Supervision}\label{sec: stage1}
We initiate the training process using a very small subset (1\%-10\%) of scientific questions annotated with ground-truth answers. The objective of this stage is not to maximize performance under supervision, but to establish stable reference anchors for both problem solving and strategy generation.\par
\subsubsection{Solver Reward}
For questions with ground-truth answers, the Solver is rewarded based on exact correctness:
\begin{equation}
r_i^{\text{sol}}=G(s_i) \in \{0, 1\}
\end{equation}
If a generated solution is consistent with the reference answer, it is regarded as correct and receives a reward of $1$; otherwise, it receives $0$. This binary reward formulation encourages the Solver to generate solutions fully consistent with the reference. This supervision does not aim to exhaustively teach domain knowledge, but instead provides a minimal alignment signal that calibrates the model's reasoning trajectories scientifically.\par

\subsubsection{Verifier Reward}
The goal of the Verifier is to generate verification strategies that are both discriminative and reliable: an optimal verification strategy should pass all correct solutions while rejecting incorrect ones.\par

First, we define the set of correct solutions as:
\begin{equation}
S^+ ( q ) = \{ s_i \in S ( q ) \mid G ( s_i ) = 1 \}
\end{equation}
If a verification strategy passes all correct solutions, it is considered positively aligned. The sign of the verification reward is defined as:
\begin{equation}
\text{sign}(v_j) =
\begin{cases}
+1, & \forall s_i \in \mathcal{S}^+(q),\; \mathrm{Eval}(s_i, v_j)=1, \\
-1, & \text{otherwise}.
\end{cases}
 \end{equation}
 The final reward function for verification is:
 \begin{equation}\label{eq:r_con}
 r_j^{\text{ver}} = \text{sign}(v_j) \cdot \mathbb{E}_{s \in \mathcal{S}^-(q)} 
\left[ 1 - \mathrm{Eval}(s, v_j) \right] 
\end{equation}
where $S^{-}(q) = S(q)\setminus S^{+}(q)$ is the set of incorrect solutions. This formulation assigns positive rewards to verification strategies that pass all correct solutions while rejecting a larger proportion of incorrect ones, and penalizes strategies that incorrectly reject correct solutions.\par

\subsubsection{Sequential Optimization}
Since the Solver and Verifier share parameters, directly optimizing both objectives jointly at this stage may lead to unstable dynamics. We therefore adopt a sequential optimization scheme after collecting rollout samples and their corresponding rewards for both solutions and strategies. Specifically, within each PPO iteration, we first update the model parameters using solution data, and then update the same parameters using strategy data. This ensures that the shared model parameters $\theta$ alternately integrate feedback signals from solving accuracy and strategy discriminability. By aligning the Solver with ground-truth correctness and training the Verifier to maximally distinguish correct solutions from incorrect ones, this stage establishes a stable and reliable foundation for the subsequent unsupervised co-evolving stage.\par
\begin{algorithm}[t]
\caption{Stage 1: Anchored Learning}
\label{alg:stage1}
\begin{algorithmic}
\STATE {\bfseries Input:} 
Sparse labeled dataset $\mathcal{D}_{\mathrm{GT}}=\{(q,a^\ast)\}$; 
shared policy $\pi_\theta$; 
number of solutions $N$; number of strategies $M$.

\STATE {\bfseries Initialize:} Policy parameters $\theta$.
\FOR{each training iteration}
    \FOR{each rollout sampled labeled question $(q,a^\ast)\in\mathcal{D}_{\mathrm{GT}}$}
        \STATE Generate $N$ solutions and $M$ verification strategies.
        \STATE Evaluate each $(s_i,v_j)$ using an external Judge Model:
        $E_{ij}=\mathrm{Eval}(s_i,v_j)\in\{0,1\}$.
        \STATE Compute solver reward using ground-truth:
        $r_i^{\mathrm{sol}} = G(s_i,a^\ast)$.
        \STATE Identify correct set:
        $\mathcal{S}^+(q)=\{s_i\mid r_i^{\mathrm{sol}}=1\}$.
        \STATE Compute verifier reward:
        
        $r_j^{\mathrm{ver}}= \text{sign}(v_j)\cdot \mathbb{E}_{s\in\mathcal{S}^-(q)} \big[1-\mathrm{Eval}(s,v_j)\big]$ 
    \ENDFOR

    \STATE Sequentially update $\theta$ using PPO with solution samples $\{(q,s_i,r_i^{\mathrm{sol}})\}$ and strategy samples $\{(q,v_j,r_j^{\mathrm{ver}})\}$.
\ENDFOR
\STATE {\bfseries Output:} Updated policy $\pi_\theta$.
\end{algorithmic}
\end{algorithm}\vspace*{-0.6\baselineskip}

%\subsection{Stage 2: Unsupervised Co-evolving without Ground-Truth}
\subsection{Unsupervised Co-evolution via Geometric Consensus.}\label{sec: stage2}
After anchored learning in Stage 1, the model acquires basic capabilities for generating candidate solutions and verification strategies. In Stage 2, we further scale training to a large corpus of unlabeled scientific questions, where no ground-truth answers are available. The key challenge in this stage is to provide reliable training signals for both the Solver and the Verifier without relying on external annotations.\par

To address this challenge, we design an unsupervised co-evolving mechanism that leverages mutual consistency between solutions and verification strategies in place of task-level answer labels. The core idea is to replace absolute correctness signals with relative agreement and structural consensus, enabling the Solver and Verifier to supervise each other.\par

\subsubsection{Solution Reward via Strategy Consensus}
Without reference answers, the Solver may reinforce incorrect reasoning through self-confirmation. To address this, we replace absolute correctness with relative consensus as the learning signal. Specifically, the quality of a solution is measured by its pass rate across various verification strategies. The solution reward is defined as:
\begin{equation}
r_i^\text{sol} = \frac{1}{M} \sum_{j=1}^{M} \text{Eval}(s_i, v_j)
\end{equation}
This relative formulation mitigates noise from individual verification errors, encouraging the model to generate solutions that remain consistent across multiple verification perspectives. In practice, we introduce a threshold $\tau$ (e.g., $\tau$=0.8) and treat solutions whose passing rate exceeds $\tau$ as high-consensus solutions. These solutions are collected into the set $\text{S}^+(q)$. The consistency-based component $r_i^{\text{cons}}$ of the verification reward, is then computed based on this set via Eq.\eqref{eq:r_con}.\par

\subsubsection{Verification Reward via Geometric Modeling}
To prevent the Verifier from maximizing consensus reward by generating homogeneous or trivial verification strategies, we propose a reward framework based on geometric modeling, evaluating verification strategies along three dimensions: consistency, reliability, and diversity.\par 

The consistency reward $r_i^{\text{cons}}$ is derived from the high-consensus solution set $\text{S}^+(q)$ via Eq.\eqref{eq:r_con}, encouraging strategies that accept high-consensus solutions while rejecting others. For the other two terms, we decouple the verification assessment from generated solutions and instead model the rewards based on the geometric structure of strategies in the latent representation space. Each natural-language verification strategy $v_j$ is mapped into a high-dimensional semantic vector $ \mathbf{z}_j = \phi(v_j) \in \mathbb{R}^d $ using a fixed pretrained embedding model Qwen3-Embedding-8B \cite{zhang2025qwen3em}. We then summarize these embeddings $\{\mathbf{z}_j\}$ by their global centroid:
\begin{equation}
\boldsymbol{\mu} = \frac{1}{M}\sum_{j=1}^{M}\mathbf{z}_j .
\end{equation}
%$\{\mathbf{z}_j\}_{j=1}^M \xrightarrow{\text{Clustering}} \{ \mathcal{C}_1, \dots, \mathcal{C}_K \}$. For each cluster $C_k$, the cluster center is $\boldsymbol{\mu}_k = \frac{1}{|\mathcal{C}_k|}\sum_{\mathbf{z}_j \in \mathcal{C}_k} \mathbf{z}_j$.

\textbf{Reliability Reward.}
Strategies that lie closer to the centroid are assumed to less likely to hallucinations or topic-drifting, representing more stable and trustworthy verification logic. We therefore define the reliability reward:
\begin{equation}\label{r_rel}
r_j^{\text{rel}} = 1 - \frac{d_j}{\max_{\ell} d_{\ell} + \epsilon}
\end{equation}
where $d_j = \| \mathbf{z}_j - \boldsymbol{\mu} \|_2$ is the Euclidean distance between verification strategy $\mathbf{z}_j$ and the centroid, and $\max_{\ell} d_{\ell}$ is the largest such distance among the $M$ strategies; strategies closer to the centroid receive higher reliability rewards.\par

\textbf{Diversity Reward.}
Furthermore, to encourage the coverage of diverse verification perspectives, we introduce a diversity reward modeled in polar coordinates. Specifically, we use Principal Component Analysis (PCA) to project the decentralized vectors $\mathbf{u}_j = \mathbf{z}_j - \boldsymbol{\mu}$ into a 2D subspace, as $\tilde{\mathbf{u}}_j = ({x}_j,{y}_j)\in \mathbb{R}^2$.
%\begin{equation}
%\mathbf{u}_j\xrightarrow{\text{PCA}}\tilde{\mathbf{u}}_j = ({x}_j,{y}_j)\in \mathbb{R}^2
%\end{equation}
Then compute the polar angle $\theta_j = \text{atan2}({x}_j,{y}_j)$ for each strategy. In the ideal state, strategies should be uniformly distributed around the centroid from a geometric viewpoint, covering a wider range of semantic directions. We promote this by rewarding samples with significant angular deviations from others, strategies that are too close to others receive lower rewards:
\begin{equation}\label{r_div}
r_j^{\text{div}} = \frac{1}{M-1} \sum_{j' \neq j} \left(1 - \cos(\theta_j -\theta_{j'})\right)
\end{equation}
The final verification reward is a weighted sum of consistency, reliability, and diversity:
\begin{equation}
r_j^{\mathrm{ver}} = 
\alpha \, r_j^{\text{con}} + \beta \, r_j^{\text{rel}} + \gamma \, r_j^{\text{div}}
\end{equation}
\par
In our experiments, we set $\alpha=1.0,\beta=0.5,\gamma=0.5$.\par

\subsubsection{Joint Optimization}
Unlike the sequential training in Anchored Learning, this stage employs a joint optimization. Specifically, solution samples and strategy samples with their respective rewards are mixed within the same training batch. Consequently, during each PPO iteration, the model parameters $\theta$ receive gradient updates simultaneously from both the Solver and Verifier tasks. This enables the model to dynamically adjust its verification criteria while exploring new solution spaces, allowing solutions and verification strategies to mutually reinforce each other and progressively improve without ground-truth supervision.\par
\begin{algorithm}[t]
\caption{Stage 2: Unsupervised Co-evolution}
\label{alg:stage2}
\begin{algorithmic}
\STATE {\bfseries Input:} 
Unlabeled dataset $\mathcal{D}_{\mathrm{U}}=\{q\}$; 
shared policy $\pi_\theta$; 
embedding model $\phi(\cdot)$;
number of solutions $N$; number of strategies $M$;
consensus threshold $\tau$.

\FOR{each training iteration}
    \FOR{each rollout sampled question $q\in\mathcal{D}_{\mathrm{U}}$}
        \STATE Generate $N$ solutions and $M$ verification strategies.
        \STATE Evaluate each $(s_i,v_j)$ using a fixed external Judge Model:
        $E_{ij}=\mathrm{Eval}(s_i,v_j)\in\{0,1\}$.
        \STATE Compute solution reward:
        $
        r_i^{\mathrm{sol}}=\frac{1}{M}\sum_{j=1}^{M}E_{ij}
        $
        \STATE Identify high-consensus solutions:
        $
        \mathcal{S}^+(q)=\{s_i\mid r_i^{\mathrm{sol}}\ge\tau\}
        $.

        \STATE Embed strategies $\mathbf{z}_j=\phi(v_j)$ and compute the global centroid $\boldsymbol{\mu}=\frac{1}{M}\sum_{j}\mathbf{z}_j$.

        \STATE Compute consistency, reliability, and diversity rewards (Eq.~\eqref{eq:r_con}, \eqref{r_rel} and \eqref{r_div}).

        \STATE Compute final verifier reward:
        $
        r_j^{\mathrm{ver}}=
        \alpha r_j^{\mathrm{con}}+
        \beta r_j^{\mathrm{rel}}+
        \gamma r_j^{\mathrm{div}}
        $
    \ENDFOR

    \STATE Update $\theta$ using PPO with mixed samples
    $\{(q,s_i,r_i^{\mathrm{sol}})\}\cup\{(q,v_j,r_j^{\mathrm{ver}})\}$.
\ENDFOR
\STATE {\bfseries Output:} Updated policy $\pi_\theta$.
\end{algorithmic}
\end{algorithm}\vspace*{-0.8\baselineskip}

\begin{table*}[t!]
\centering
\small
\setlength{\tabcolsep}{1.8mm}
\caption{\textbf{Training Data Composition of Different Scales.} 
The second column represents the subset composition of disciplines in MegaScience.}
\label{tab:data_composition}
\begin{tabular}{c c c c}
\toprule
\textbf{Scale} & \textbf{MegaScience} & \textbf{NuminaMath} & \textbf{Other} \\
\midrule
4k & Phy 1k, Bio 1k, Chem 1k & 1k & -- \\
18k & Phy 5k, Bio 2k, Chem 2k, Med 1k, Math 1k, CS 1k, Eco 1k & 5k & -- \\
30k & Phy 5k, Bio 4k, Chem 4k, Med 4k, Math 2k, CS 2k, Eco 2k & 5k & ScienceQA 1k, CaseHold 1k \\
\bottomrule
\end{tabular}
\vspace{-4pt}
\end{table*}
\section{Theoretical Analysis}

As mentioned above, Sci-CoE has shown expressive effectiveness in the training process. Here, we discuss its deeper theoretical factors as follows. \par

To rigorously justify the necessity of the proposed Geometric Reward, we provide a theoretical analysis of the co-evolution dynamics between the Solver $\pi_S$ and the Verifier $\pi_V$. We define the verification process as a mapping where a strategy $v \in \mathcal{V}$ evaluates a solution $s \in \mathcal{S}$, yielding a binary score $\mathbb{I}(s \vdash v) \in \{0, 1\}$.

%\subsection{The Mode Collapse of Naive Reward}
\subsection{Why a Naive Consensus Reward Can Degenerate?}
In a naive self-play paradigm, the Verifier is optimized solely based on the consensus reward, which can be simplified as:
\begin{equation}
\max_{\pi_V} \mathcal{J}_{\text{naive}}(\pi_V) = \mathbb{E}_{s \sim \pi_S,\, v \sim \pi_V} [\mathbb{I}(s \vdash v)]
\end{equation}

During the initial stages of training, the generated strategies are highly diverse. However, since a negative reward is incurred whenever a single correct solution is missed, the Verifier can exploit conservative, superficial, or tautological verification strategies (e.g., trivial format checks) that make $\mathbb{I}(s \vdash v) = 1$ with near certainty, regardless of the intrinsic scientific correctness of $s$. \par
Let $\mathcal{H}(\pi_V)$ denote the Shannon entropy of the Verifier's strategy distribution. The key point is that the naive objective admits such a degenerate solution: there exist near-trivial policies with low verification strategy entropy $\mathcal{H}(\pi_V)$ and low solution-verification mutual information $\mathcal{I}(S; V)$ that still attain near-maximal consensus reward. Whenever the Verifier drifts toward this regime, it loses discriminative power, and the Solver receives dense but uninformative gradients that stall the self-improvement loop. This failure mode is what the geometric reward is designed to prevent.

%\subsection{Information-Theoretic Perspective of Geometric Reward}
\subsection{How the Geometric Reward Counteracts Degeneration?}

Our geometric reward addresses this degeneration by embedding discrete textual strategies into a continuous latent semantic space $\mathcal{Z}$ via a fixed map $\phi: \mathcal{V} \to \mathcal{Z}$, and augmenting the objective with two geometric regularizers:
\begin{equation}
\mathcal{J}_{\text{geo}}(\pi_V) = \alpha \mathbb{E}[R_{\text{con}}] + \beta \mathbb{E}[R_{\text{rel}}(Z)] + \gamma \mathbb{E}[R_{\text{div}}(Z)]
\end{equation}

\textbf{Reliability ($R_{\text{rel}}$):} By penalizing the Euclidean distance to the semantic centroid $\boldsymbol{\mu}$, $R_{\text{rel}}$ acts as a prior regularization. It encourages the sampled strategy $z_i$ to remain within the high-density region of the strategy distribution, discouraging the Verifier from drifting toward off-topic or nonsensical text merely to inflate the diversity term.\par
\textbf{Diversity ($R_{\text{div}}$):} The angular distance maximization in the polar coordinate space encourages the strategies $\{z_1, z_2, \ldots\}$ to occupy different semantic directions around the centroid. Increasing pairwise angular distance spreads the strategies over a wider range of directions in the projected space; we use this as a tractable geometric surrogate for semantic diversity.\par

Consequently, the consistency, reliability, and diversity terms provide complementary inductive biases: consistency ties rewards to solution quality, reliability keeps strategies prototypical, and diversity discourages homogeneous verification. The intended effect is to encourage the Verifier to construct multi-dimensional evaluation criteria (e.g., dimensional analysis, extreme-value testing) so that it supplies a more stable and less collapse-prone learning signal for the Solver's evolution.

\subsection{Effect of Verifier Collapse on the Solver Signal}
Verifier collapse directly weakens the learning signal for the Solver. Since $r_i^{\mathrm{sol}}=\frac{1}{M}\sum_{j=1}^{M}E_{ij}$,
if all verification strategies become trivial and accept every solution, then $E_{ij}=1$ for all $i,j$, so every candidate receives the same reward. Although the average reward is high, it cannot distinguish correct reasoning trajectories from incorrect ones. Maintaining diverse and discriminative verification strategies is therefore necessary to preserve informative Solver rewards.
The effectiveness of consensus depends on both the accuracy and the correlation of verification strategies. If each strategy has error rate $\epsilon<1/2$ and their pairwise error correlation is $\rho$, then the variance of the average judgment satisfies
\begin{equation}
\operatorname{Var}(\bar X)
=
\frac{\epsilon(1-\epsilon)}{M}\left[1+(M-1)\rho\right].
\end{equation}
Increasing the number of strategies reduces this variance appreciably only when their errors are not highly correlated ($\rho$ small); as $\rho \to 1$ the variance approaches $\epsilon(1-\epsilon)$ regardless of $M$. This provides a concrete motivation for jointly encouraging strategy reliability and diversity rather than merely increasing the number of strategies.

\section{Experiments}

\subsection{Experimental Setup}
\textbf{Model and Optimization Configuration.}
We employ Qwen2.5-7B-Instruct, Qwen3-8B \cite{yang2025qwen3} and Qwen3.5-4B~\cite{qwen3.5} as the base policy models to concurrently perform Solver and Verifier tasks. We additionally include Agents-A1-4B \cite{bai2026scalinghorizonparametersreaching} as a 4B-scale agentic policy model to examine whether Sci-CoE generalizes to an  agentic model backbone. To provide high-quality verification feedback signals during training, that is to judge whether a Solver's solution passes the Verifier's generated strategy, we utilize Qwen3-235B-A22B \cite{yang2025qwen3} as the fixed external judging model.\par

%At each sampling step during reinforcement learning, we generate 10 rollouts for solutions and 10 for verification strategies using vLLM \cite{kwon2023efficient} with a temperature of 1.0, top-p of 1.0. For optimization, we set the learning rate to $1 \times 10^{-6}$ and the KL coefficient $\beta$ to 0.01.\par

\noindent \textbf{Data Construction.}
We integrate datasets including MegaScience \cite{fan2025megascience}, Numinamath \cite{li2024numinamath}, ScienceQA \cite{saikh2022scienceqa}, and CaseHold \cite{zheng2021casehold}, covering diverse scientific domains such as mathematics, physics, chemistry, and biology. In the Anchored Learning Stage, we sample 4k data from MegaScience and Numinamath annotated with ground-truth reference answers. In the Unsupervised Co-evolution Stage, we further construct two unlabeled training sets of different scales of 18k and 30k to test scalability. Detailed dataset compositions are provided in \ref{tab:data_composition}.\par

\noindent \textbf{Benchmarks.}
To test scientific reasoning ability, we evaluate our approach on the following benchmarks: MMLU-Pro \cite{wang2024mmlu}, GPQA-Diamond \cite{rein2024gpqa}, and UGPhysics \cite{xu2025ugphysics}. These benchmarks collectively cover a wide range of scientific disciplines with challenging tasks, offering a stricter evaluation of complex reasoning abilities. For evaluation, we strictly employ the official evaluation scripts provided for each dataset to ensure accuracy and comparability.\par

\subsection{Main Results}
\begin{figure}[t]
\vspace{-6pt}
    \centering
    \includegraphics[width=\columnwidth]{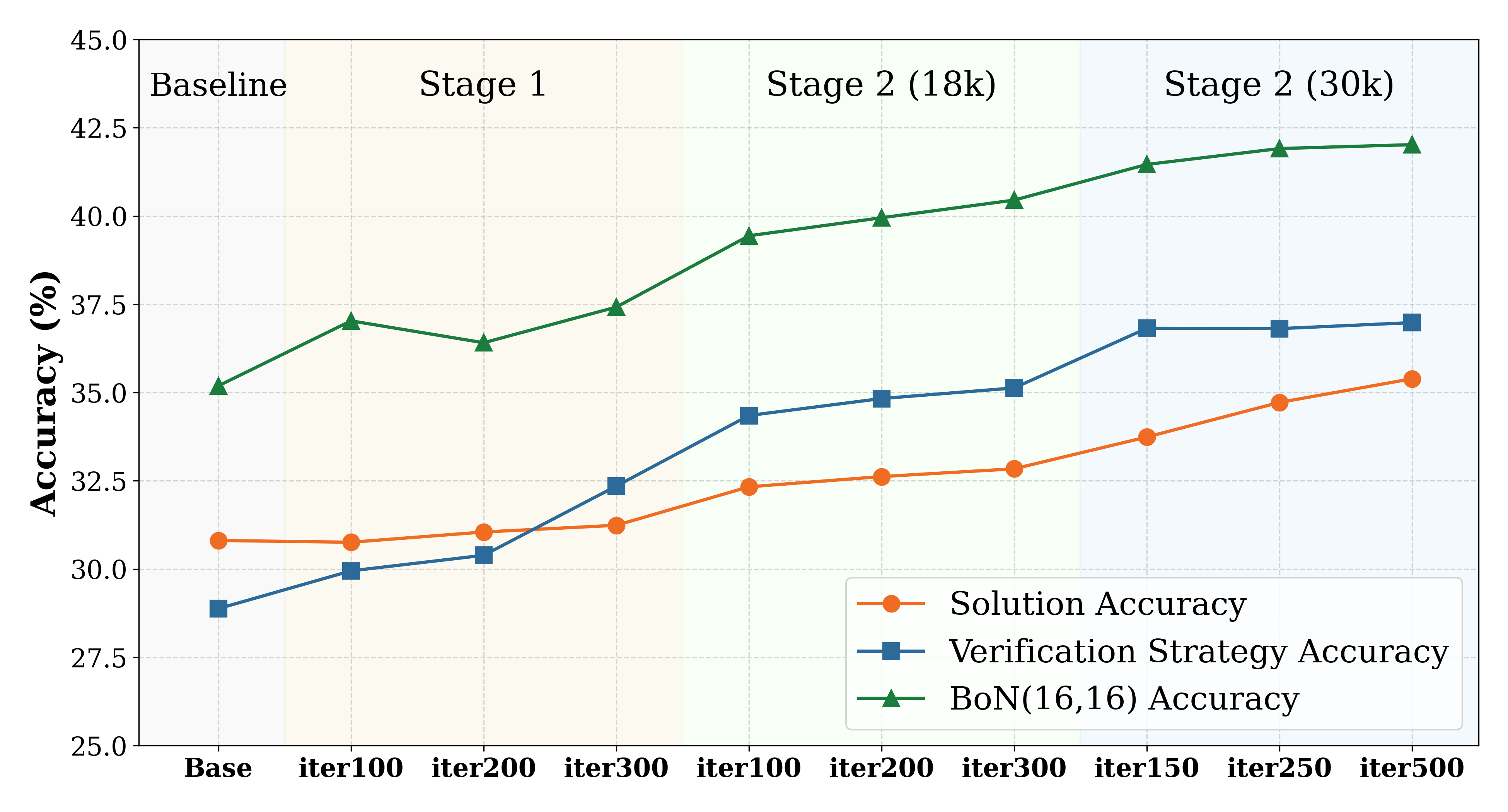}
\caption{Performance of the model at different stages of the training process on GPQA-D evaluation data, the three broken lines represent the average accuracy of the generated solutions, the average accuracy of the generated validation strategies and Best-of-N(BoN) accuracy, using 16 generated solutions and 16 generated strategies. The baseline is Qwen2.5-7B-Instruct, and the rest of the model names represent the number of training steps in that stage.}
\label{fig:train_trend}
\vspace{-6pt}
\end{figure}

\begin{table*}[t]
\centering
\small
\setlength{\tabcolsep}{3pt}
\caption{\textbf{Main Results on MMLU-Pro.} We report the accuracy (\%) on MMLU-Pro and its subsets. The best results within each model block and benchmark are highlighted in \textbf{bold}, and \underline{underline} indicates the second best.}
\label{tab:mmlu_results}
\begin{tabular}{@{}l>{\columncolor{gray!15}}ccccccccccccccc@{}}
\toprule
\textbf{Model} & \textbf{Overall} & \textbf{\textit{Bio.}} & \textbf{\textit{Bus.}} & \textbf{\textit{Che.}} & \textbf{\textit{C.S.}} & \textbf{\textit{Eco.}} & \textbf{\textit{Eng.}} & \textbf{\textit{Hea.}} & \textbf{\textit{His.}} & \textbf{\textit{Law}} & \textbf{\textit{Math}} & \textbf{\textit{Phi.}} & \textbf{\textit{Phy.}} & \textbf{\textit{Psy.}} & \textbf{\textit{Oth.}} \\
\midrule
\multicolumn{16}{@{}l}{\textit{Other Models}} \\
\quad Llama-3.1-8B-Instruct & 44.25 & 63.04 & 49.30 & 37.63 & 48.29 & 55.09 & 29.72 & 50.73 & 42.26 & 27.25 & 43.82 & 44.49 & 40.26 & 60.03 & 44.81 \\
%\quad DeepSeekMath-7B-Instruct & 35.30 & 46.00 & 42.33 & 41.08 & 39.02 & 48.22 & 33.64 & 25.06 & 15.22 & 15.71 & 42.78 & 27.05 & 39.18 & 39.47 & 28.03 \\
\quad Ministral-8B-Instruct & 37.93 & 59.00 & 39.42 & 26.41 & 44.88 & 49.29 & 23.12 & 43.28 & 43.83 & 25.98 & 41.15 & 36.47 & 29.18 & 51.63 & 40.15 \\
\quad Mathstral-7B-v0.1 & 42.00 & 63.46 & 42.08 & 38.78 & 45.61 & 51.78 & 38.39 & 39.24 & 35.96 & 22.43 & 47.67 & 38.28 & 38.03 & 52.63 & 40.91 \\
\quad Yi-1.5-9B-Chat & 45.95 & 66.67 & 54.25 & 39.49 & 50.00 & 60.19 & 33.23 & 43.52 & 40.94 & 26.61 & 52.48 & 40.08 & 41.42 & 59.40 & 44.91 \\
\quad Mistral-Small-Instruct & 48.40 & 71.69 & 52.72 & 36.84 & 53.66 & 60.07 & 30.55 & 53.79 & 50.13 & 31.97 & 50.85 & 48.10 & 40.34 & 63.91 & 55.09 \\
\quad GPT-4o-2024-08-06 & 74.68 & 89.26 & 80.10 & 72.70 & 78.29 & 81.64 & 55.31 & 76.04 & 73.23 & 58.95 & 79.42 & 79.55 & 70.34 & 75.06 & 82.71 \\
\quad Deepseek-v3 & 75.87 & 86.89 & 80.99 & 79.51 & 79.51 & 81.75 & 62.23 & 73.72 & 68.24 & 54.77 & 86.16 & 76.41 & 71.54 & 78.98 & 78.82 \\
\midrule
\multicolumn{16}{@{}l}{\textit{Qwen2.5-7B-Instruct}} \\
\quad Base Model & 57.39 & 72.11 & 64.89 & \textbf{57.16} & 60.49 & \underline{68.84} & 39.94 & 56.85 & 48.29 & 32.52 & 71.87 & \underline{49.90} & 58.35 & 65.91 & 54.33 \\
\quad Sci-CoE-Stage 1 & 57.68 & \textbf{74.34} & 67.17 & \underline{56.89} & 60.73 & 68.72 & \underline{40.04} & 56.60 & 47.77 & \underline{33.33} & 72.09 & 48.50 & \underline{59.05} & 65.54 & 53.90 \\
\quad Sci-CoE-Stage 2-18k & \underline{58.05} & 73.50 & \textbf{68.69} & \textbf{57.16} & \underline{61.22} & 68.01 & \underline{40.04} & \underline{58.07} & \textbf{49.08} & 32.61 & \underline{72.54} & \textbf{50.50} & 58.97 & \underline{66.67} & \underline{54.65} \\
\quad Sci-CoE-Stage 2-30k & \textbf{58.51} & \underline{73.92} & \underline{68.19} & 55.39 & \textbf{61.71} & \textbf{70.62} & \textbf{42.31} & \textbf{58.19} & \underline{48.82} & \textbf{34.06} & \textbf{72.76} & \textbf{50.50} & \textbf{59.35} & \textbf{67.17} & \textbf{54.87} \\
\midrule
\multicolumn{16}{@{}l}{\textit{Qwen3-8B}} \\
\quad Base Model & 63.19 & 78.80 & 69.71 & 68.02 & 66.10 & 72.27 & 53.04 & 62.47 & 51.97 & 31.52 & 78.53 & 51.50 & 67.67 & 69.30 & \underline{55.95} \\
\quad Sci-CoE-Stage 1 & 63.27 & 78.94 & 69.20 & 66.78 & 65.85 & 72.63 & \underline{53.77} & \underline{63.08} & 50.92 & \underline{32.61} & 78.09 & 52.51 & \textbf{68.44} & \underline{69.42} & 55.41 \\
\quad Sci-CoE-Stage 2-18k & \underline{63.56} & \underline{79.22} & \textbf{70.85} & \underline{68.02} & \underline{66.34} & \underline{72.87} & 53.35 & 62.71 & \underline{51.44} & 32.52 & \underline{79.42} & \underline{52.91} & 67.74 & 69.17 & 55.19 \\
\quad Sci-CoE-Stage 2-30k & \textbf{64.34} & \textbf{80.20} & \underline{70.72} & \textbf{68.20} & \textbf{68.05} & \textbf{73.93} & \textbf{54.59} & \textbf{63.33} & \textbf{54.07} & \textbf{33.42} & \textbf{79.79} & \textbf{53.51} & \underline{68.36} & \textbf{70.30} & \textbf{56.06} \\
\midrule
\multicolumn{16}{@{}l}{\textit{Qwen3.5-4B}} \\
\quad Base Model & 78.25 & 89.12 & 83.14 & 88.60 & 80.24 & \underline{83.65} & 75.13 & 73.35 & 62.47 & 47.96 & \textbf{93.78} & \underline{65.33} & 88.99 & 77.57 & 67.10 \\
\quad Sci-CoE-Stage 1 & 78.28 & 88.98 & 83.52 & \textbf{89.22} & \underline{81.71} & 83.53 & \underline{75.23} & \underline{73.59} & 62.73 & 47.87 & 93.26 & 65.13 & \underline{89.07} & 76.69 & 67.21 \\
\quad Sci-CoE-Stage 2-18k & \underline{78.38} & \textbf{90.66} & \underline{83.90} & 88.07 & \textbf{82.20} & 82.35 & 72.76 & 73.47 & \underline{63.52} & \underline{50.50} & \underline{93.41} & 64.13 & 88.61 & \underline{77.82} & \underline{68.40} \\
\quad Sci-CoE-Stage 2-30k & \textbf{79.80} & \underline{90.10} & \textbf{84.54} & \underline{88.69} & 81.46 & \textbf{84.72} & \textbf{76.16} & \textbf{74.21} & \textbf{63.78} & \textbf{53.13} & 93.26 & \textbf{69.54} & \textbf{89.61} & \textbf{79.95} & \textbf{70.78} \\
\midrule
\multicolumn{16}{@{}l}{\textit{Agents-A1-4B}} \\
\quad Base Model & 77.28 & 90.24 & 82.76 & 87.54 & 80.00 & 81.75 & 69.76 & 74.33 & 61.15 & 48.32 & 92.89 & 62.73 & 88.45 & 75.94 & 66.77 \\
\quad Sci-CoE-Stage 1 & 77.77 & 89.68 & 83.40 & \textbf{88.16} & 80.73 & 82.35 & 72.14 & \underline{74.45} & \underline{62.47} & 48.05 & \underline{93.63} & 63.73 & 88.53 & 76.19 & 66.67 \\
\quad Sci-CoE-Stage 2-18k & \underline{78.17} & \underline{91.07} & \underline{83.53} & 87.90 & \underline{81.95} & \textbf{82.70} & \underline{72.24} & 74.21 & 61.68 & \underline{48.86} & 93.56 & \underline{64.53} & \underline{88.76} & \textbf{77.69} & \underline{67.64} \\
\quad Sci-CoE-Stage 2-30k & \textbf{78.72} & \textbf{91.35} & \textbf{84.41} & \underline{88.07} & \textbf{82.44} & \underline{82.46} & \textbf{72.45} & \textbf{75.55} & \textbf{62.73} & \textbf{50.14} & \textbf{93.86} & \textbf{66.33} & \textbf{89.53} & \underline{77.07} & \textbf{68.29} \\

\bottomrule
\end{tabular}
\end{table*}

\begin{table*}[t]
\centering
\small
\caption{\textbf{Main Results on UGPhysics.} We report the accuracy (\%) on the English subset of UGPhysics. The best results within each model block and benchmark are highlighted in \textbf{bold}, and \underline{underline} indicates the second best. In case of ties, all tied results are marked. ``Mec.'', ``Elec.'' and ``Modern'' stand for Mechanics \& Thermodynamics, Electromagnetism, and Modern Physics subsets of UGPhysics. We use Qwen3-235B-A22B as the judging model.}
\label{tab:ugp_results}
\begin{tabular}{@{}lc>{\columncolor{gray!15}}cccc@{}}
\toprule
\textbf{Model} & \textbf{Data Scale} & \textbf{Overall Acc} & \textbf{\textit{Mec. and Ther.}} & \textbf{\textit{Elec.}} & \textbf{\textit{Modern Physics}} \\
\midrule
\multicolumn{6}{@{}l}{\textit{Other Models}} \\
\quad Llama-3.1-8B-Instruct & --  & 14.66 & 12.64 & 14.35 & 16.80 \\
%\quad DeepSeekMath-7B-Instruct & -- & 16.21 & 13.25 & 16.49 & 19.07 \\
\quad Ministral-8B-Instruct-2410 & -- & 16.39 & 13.95 & 15.52 & 19.20 \\
\quad Mathstral-7B-v0.1 & -- & 17.45 & 14.82 & 17.77 & 19.94 \\
\quad Yi-1.5-9B-Chat & -- & 17.61 & 16.00 & 15.85 & 19.94 \\
\quad Mistral-Small-Instruct-2409 & -- & 25.72 & 22.71 & 22.70 & 29.97\\
%\quad DeepSeek-R1-Distill-Qwen-7B & -- & 29.17 & 29.25 & 27.41 & 29.80\\
\quad Phi-4 & -- & 37.16 & 34.13 & 36.51 & 40.45\\
\quad GPT-4o-2024-08-06 & -- & 39.29 & 36.97 & 36.40 & 42.80\\
\midrule
\multicolumn{6}{@{}l}{\textit{Qwen2.5-7B-Instruct}} \\
\quad Base Model & -- & 20.67 & 18.88 & 18.52 & \underline{23.34} \\
\quad Sci-CoE-Stage 1 & 4k & 21.07 & 20.14 & 19.81 & 22.51 \\
\quad Sci-CoE-Stage 2 & 18k & \underline{21.92} & \underline{20.92} & \underline{21.31} & 23.17 \\
\quad Sci-CoE-Stage 2 & 30k & \textbf{22.64} & \textbf{21.84} & \textbf{23.13} & \textbf{24.91} \\
\midrule
\multicolumn{6}{@{}l}{\textit{Qwen3-8B}} \\
\quad Base Model & -- & 31.76 & \textbf{30.73} & 29.98 & 33.51 \\
\quad Sci-CoE-Stage 1 & 4k & 32.03 & 30.25 & 30.62 & \underline{34.38} \\
\quad Sci-CoE-Stage 2 & 18k & \underline{32.46} & 30.21 & \underline{33.30} & \underline{34.38} \\
\quad Sci-CoE-Stage 2 & 30k & \textbf{33.10} & \underline{30.51} & \textbf{34.80} & \textbf{34.99} \\
\midrule
\multicolumn{6}{@{}l}{\textit{Qwen3.5-4B}} \\
\quad Base Model & -- & 43.57 & \textbf{45.64} & 38.76 & \underline{43.46} \\
\quad Sci-CoE-Stage 1 & 4k & 43.35 & 44.51 & 39.40 & \textbf{43.80} \\
\quad Sci-CoE-Stage 2 & 18k & \underline{43.61} & \underline{45.51} & 40.15 & 43.11 \\
\quad Sci-CoE-Stage 2 & 30k & \textbf{43.70} & 45.47 & \textbf{40.47} & 43.24 \\
\bottomrule
\end{tabular}
\end{table*}

\begin{table*}[t]
\centering
\small
\caption{\textbf{Main Results on GPQA-Diamond.} We report the accuracy (\%) on GPQA-Diamond and its subsets. The best results within each model block and benchmark are highlighted in \textbf{bold}, and \underline{underline} indicates the second best.}
\label{tab:gpqa_results}
\begin{tabular}{@{}lc>{\columncolor{gray!15}}cccc@{}}
\toprule
\textbf{Model} & \textbf{Data Scale} & \textbf{Overall Acc} & \textbf{\textit{Physics}} & \textbf{\textit{Chemistry}} & \textbf{\textit{Biology}} \\
\midrule
\multicolumn{6}{@{}l}{\textit{Qwen2.5-7B-Instruct}} \\
\quad Base Model & -- & 30.81 & 33.73 & \underline{24.73} & \textbf{47.37} \\
\quad Sci-CoE-Stage 1 & 4k & 31.31 & \underline{34.88} & \underline{24.73} & \textbf{47.37} \\
\quad Sci-CoE-Stage 2 & 18k & \underline{33.33} & \textbf{41.86} & 23.66 & \underline{42.11} \\
\quad Sci-CoE-Stage 2 & 30k & \textbf{35.35} & \textbf{41.86} & \textbf{26.88} & \textbf{47.37} \\
\midrule
\multicolumn{6}{@{}l}{\textit{Qwen3-8B}} \\
\quad Base Model & -- & 61.62 & 82.56 & 41.94 & 63.16 \\
\quad Sci-CoE-Stage 1 & 4k & 62.63 & \textbf{84.88} & 39.78 & 73.68 \\
\quad Sci-CoE-Stage 2 & 18k & \underline{64.14} & \underline{83.72} & \underline{41.94} & \underline{84.21} \\
\quad Sci-CoE-Stage 2 & 30k & \textbf{65.66} & \underline{83.72} & \textbf{44.09} & \textbf{89.47} \\
\midrule
\multicolumn{6}{@{}l}{\textit{Qwen3.5-4B}} \\
\quad Base Model & -- & 71.21 & \textbf{88.37} & 56.99 & 63.16 \\
\quad Sci-CoE-Stage 1 & 4k & \underline{74.24} & 86.05 & \textbf{67.74} & 52.63 \\
\quad Sci-CoE-Stage 2 & 18k & 73.74 & \underline{87.21} & 62.37 & \underline{68.24} \\
\quad Sci-CoE-Stage 2 & 30k & \textbf{74.75} & \textbf{88.37} & \underline{63.44} & \textbf{68.42} \\
\bottomrule
\end{tabular}
\end{table*}
\noindent \textbf{Performance on Scientific Reasoning Benchmarks.}
As shown in Table \ref{tab:mmlu_results}, Table \ref{tab:ugp_results} and Table \ref{tab:gpqa_results}, Sci-CoE consistently outperforms the corresponding base models on both general and domain-specific reasoning benchmarks. To highlight, our framework Sci-CoE with the Qwen3-8B outperforms the baseline by \textbf{4.04\%} on GPQA-Diamond dataset, raising the accuracy from 61.62 to 65.66. On the larger and broader MMLU-Pro benchmark, Sci-CoE also achieves \textbf{1.15\%} improvement, from 63.19 to 64.34, demonstrating the general applicability of the learned reasoning and verification capabilities across scientific domains. \par

For Agents-A1-4B, Sci-CoE also yields progressive gains on MMLU-Pro. The final model improves over the base model by \textbf{1.44\%} and achieves gains across all fourteen subject subsets, with particularly notable improvements in philosophy (+3.60\%), engineering (+2.69\%), and computer science (+2.44\%), further demonstrating the robustness and scalability of Sci-CoE across model backbones. \par

For UGPhysics which focuses on undergraduate-level physics reasoning, Sci-CoE does not exhibit monotonic improvements across all sub-disciplines, likely due to domain distribution mismatches between training data and specific physics topics. Nevertheless, Sci-CoE consistently improves the overall average accuracy, achieving increases of \textbf{1.97\%} and \textbf{1.34\%} respectively on the 7B and 8B base model. This suggests that Sci-CoE primarily learns general reasoning and verification patterns, rather than overfitting to specific subfields, enabling stable performance gains even when domain-specific supervision is extremely sparse.\par
We compare Sci-CoE with several baselines of other models, including Llama-3.1-8B-Instruct~\cite{dubey2024llama}, Ministral-8B-Instruct-2410~\cite{ministral8b}, Mathstral-7B-v0.1~\cite{mathstral2023}, Yi-1.5-9B-Chat~\cite{young2024yi}, Mistral-Small-Instruct-2409~\cite{jiang2023mistral7b}, as well as Phi-4~\cite{abdin2024phi}, GPT-4o~\cite{hurst2024gpt} and Deepseek-v3~\cite{liu2024deepseek}. As shown in in Table \ref{tab:mmlu_results} and Table \ref{tab:ugp_results}, Sci-CoE consistently outperforms all same-scale baselines on both the general reasoning benchmark MMLU-Pro and the domain-specific benchmark UGPhysics.

\begin{figure*}[t]
\vspace{-6pt}
    \centering
    \includegraphics[width=0.95\linewidth]{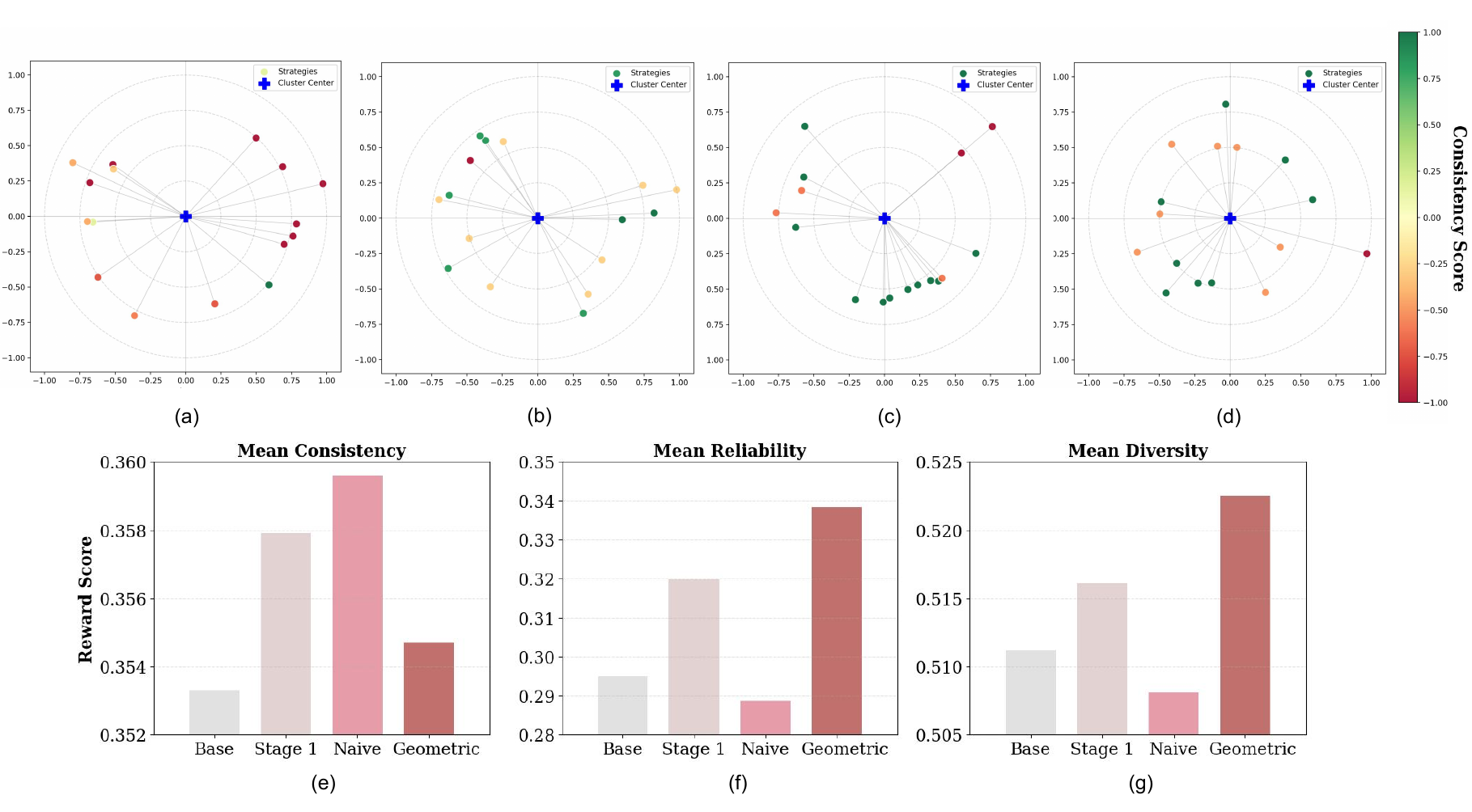}
\vspace{-6pt}
\caption{Visualization of geometric reward and quantitative analysis of verification strategies. (a)-(d) display PCA projections of strategy embeddings in a polar coordinate system across different training stages, respectively corresponding to Baseline Model, Stage 1 only, Stage 2 with Naive Consensus Reward, and Stage 2 with Geometric Reward. The angular distribution of points indicates diversity, while the radial distance to the cluster center represents strategy reliability, with closer points indicating more stable. And the color represents the consistency score, with closer to green indicating higher scores. (e)-(g) illustrate mean consistency, reliability, and diversity reward scores of different models.}
\label{fig:cluster}
\end{figure*}

\noindent \textbf{Scalability on Unlabeled Data.}
After introducing large-scale unlabeled data, Sci-CoE demonstrates strong scalability. As the scale of unlabeled data in Stage 2 increases from 18k to 30k, we observe continuous improvements in reasoning accuracy, without evident performance saturation. This indicates that increasing the diversity and quantity of unlabeled scientific problems enables Sci-CoE to discover more robust reasoning patterns. Crucially, our framework effectively bypasses the performance plateau commonly encountered in self-training, maintaining a strong Scaling Law in the absence of ground-truth supervision.\par

\noindent \textbf{Evolutionary Trends and Co-evolving Iterations.}
The performance improvements achieved by Sci-CoE are progressive, as illustrated in Figure \ref{fig:train_trend}, reflecting a stable and promising co-evolutionary process. The final model substantially outperforms the baseline in solution, verification strategy, and Best-of-N (BoN) performance. In Stage 1, the rapid improvement in verification strategy accuracy equips the model with initial evaluation capabilities for reasoning processes, providing higher-quality feedback for the Solver in subsequent unsupervised co-evolution.\par

\begin{table*}[t]
\centering
\small
\caption{\textbf{Results on Verification Capability Benchmarks.} We report the results on VerifyBench-Hard and SCI-VerifyBench. The best results within each column are highlighted in \textbf{bold}, and \underline{underline} indicates the second best.}
\label{tab:verify_results}
\begin{tabular}{@{}l>{\columncolor{gray!15}}ccccc>{\columncolor{gray!15}}cccccc@{}}
\toprule
 & \textbf{VB-Hard} & Num & Exp & MC & Str & \textbf{SCI-VB} & Math & Physics & Chemistry & Biology & QA \\
\midrule
\quad \textbf{Qwen3-8B} & 75.20 & \underline{70.24} & \underline{76.14} & 81.16 & 69.13 & 82.36 & \underline{87.00} & 76.20 & 81.40 & \textbf{85.80} & 81.40 \\
\quad Sci-CoE-Stage 1-4k & \underline{75.90} & \textbf{71.43} & 73.86 & \underline{81.86} & \underline{70.43} & 82.48 & \underline{87.00} & \underline{76.80} & \textbf{83.40} & 83.00 & 82.20 \\
\quad Sci-CoE-Stage 2-18k & 75.80 & 69.44 & \textbf{80.68} & \textbf{82.33} & 68.70 & \underline{82.60} & \textbf{87.20} & \underline{76.80} & 82.40 & 83.80 & \underline{82.80} \\
\quad Sci-CoE-Stage 2-30k & \textbf{76.10} & \underline{70.24} & \underline{76.14} & 81.63 & \textbf{72.17} & \textbf{83.36} & 86.80 & \textbf{79.20} & \underline{83.20} & \underline{84.40} & \textbf{83.20}\\
\bottomrule
\end{tabular}
\end{table*}

\begin{table*}[t]
\centering
\small
\caption{\textbf{Results on Reward Model Benchmark RM-Bench and its subsets.} The best results within each column are highlighted in \textbf{bold}, and \underline{underline} indicates the second best.}
\label{tab:rm_results}
\begin{tabular}{@{}lc>{\columncolor{gray!15}}cccccccc@{}}
\toprule
 & \textbf{Data Scale} & \textbf{RM-Bench} & Chat & Code & Math & Safety & Easy & Normal & Hard \\
\midrule
\quad \textbf{Qwen3-8B} & -- & 83.90 & 78.29 & 71.83 & 91.28 & \textbf{94.18} & 90.32 & 86.65 & 74.72\\
\quad Sci-CoE-Stage 1 & 4k & \underline{84.49} & \textbf{81.31} & 71.49 & \underline{91.47} & 93.68 & 90.50 & \textbf{86.83} & \textbf{76.12}\\
\quad Sci-CoE-Stage 2 & 18k & 84.34 & 80.28 & 71.49 & \textbf{91.79} & 93.80 & 90.90 & 86.36 & 75.76\\
\quad Sci-CoE-Stage 2 & 30k & \textbf{84.50} & \underline{80.79} & \underline{71.78} & 91.45 & \underline{93.98} & \textbf{90.91} & \underline{86.68} & \underline{75.91}\\
\bottomrule
\end{tabular}
\end{table*}

\begin{table*}[ht]
\caption{Ablation study on Scientific Reasoning benchmarks.
}
\centering
\small
\setlength{\tabcolsep}{2.3 mm}
\vspace{-4pt}
\begin{tabular}{cccccccc}
\toprule
Index&
 \makecell{Anchored \\ Learning} & \makecell{Stage1 \\ Data Scale} & \makecell{Geometric\\Reward} & \makecell{GPQA-D} & \makecell{MMLU-Pro} & \makecell{UGPhysics} & \makecell{Avg Acc}\\
\midrule
Baseline & -- & -- & -- & 61.62 &63.19 &31.76 &43.94\\ %baseline
1 &\ding{55} & \ding{55} & \ding{55} & 58.08 &63.00 &31.61 &43.49\\ %zero-gt-18k
2 &\ding{51} & 0.4k & \ding{55} & 62.12 &63.36 &31.79 & 44.17\\ %0.4k-18k
3 &\ding{51} & 4k & \ding{55} & 63.13 &63.27 &32.07 &44.41\\ %naive-18k
4 &\ding{51} & 4k & \ding{51} & \textbf{64.14} & \textbf{63.53} &\textbf{32.46} &\textbf{44.96}\\ %geometric-18k
\bottomrule
\label{tab:ablation_result}
\end{tabular}
\vspace{-4pt}
\end{table*}

\begin{table*}[t]
\centering
\small
\caption{\textbf{Ablation Study on the Source of Verification Signals.} We compare different reward mechanisms using the Qwen3-8B backbone. Direct-Judge Model relies on a 235B model's zero-shot evaluation without strategies. Self-Judge Model uses the policy model itself to execute the generated strategies. Both lead to performance degradation, highlighting the necessity of our decoupled strategy-execution paradigm.}
\setlength{\tabcolsep}{1.5 mm}
\begin{tabular}{l|c|cccc}
\toprule
\textbf{Model} & \textbf{\makecell{Verification \\ Strategy}} & \textbf{GPQA-D} & \textbf{MMLU-Pro} & \textbf{UGPhysics} & \textbf{Avg Acc} \\
\midrule
Base Model & - & 61.62 & 63.19 & 31.76 & 43.94 \\
\midrule
\rowcolor{gray!10} \textit{Alternative Reward Signals} & & & & & \\
Direct-Judge Model & \ding{55} & 58.08 & 62.98 & 31.25 & 42.50 \\
Self-Judge Model & \ding{51} & 59.60 & 63.16 & 31.45 & 42.74 \\
\midrule
\rowcolor{gray!10} \textit{Ours} & & & & & \\
Sci-CoE & \ding{51} & \textbf{64.14} & \textbf{63.53} & \textbf{32.46} & \textbf{44.96} \\
\bottomrule
\label{tab:judge_ablation}
\end{tabular}
\vspace{-0.3cm}
\end{table*}

\begin{table}[htbp]
\centering
\small
\setlength{\tabcolsep}{3pt}
\caption{\textbf{Results under different scale models.} We report the results on three benchmarks using different-sized models in the Qwen3.5 series. The best results within each model block and benchmark are highlighted in \textbf{bold}.}
\label{tab:3.5_series_results}
\begin{tabular}{@{}lcccc@{}}
\toprule
 & \textbf{Data} & \textbf{GPQA-D} & \textbf{MMLU-Pro} & \textbf{UGPhysics} \\
\midrule
\quad \textbf{Qwen3.5-9B} & -- & \textbf{81.31} & 81.06 & 45.07 \\
\quad Sci-CoE-Stage 1 & 4k & \textbf{81.31} & 81.27 & 44.38 \\
\quad Sci-CoE-Stage 2 & 18k & 79.29 & \textbf{81.34} & \textbf{45.96} \\
\midrule
\quad \textbf{Qwen3.5-4B} & -- & 71.21 & 78.25 & 43.57 \\
\quad Sci-CoE-Stage 1 & 4k & \textbf{74.24} & 78.28 & 43.35 \\
\quad Sci-CoE-Stage 2 & 18k & 73.74 & \textbf{78.38} & \textbf{43.61} \\
\midrule
\quad \textbf{Qwen3.5-2B} & -- & 52.02 & 62.92 & 18.91 \\
\quad Sci-CoE-Stage 1 & 4k & 54.04 & \textbf{63.61} & 19.17 \\
\quad Sci-CoE-Stage 2 & 18k & \textbf{55.05} & 63.40 & \textbf{19.35} \\
\midrule
\quad \textbf{Qwen3.5-0.8B} & -- & 35.35 & \textbf{40.91} & 7.59 \\
\quad Sci-CoE-Stage 1 & 4k & 31.82 & 40.77 & 7.57 \\
\quad Sci-CoE-Stage 2 & 18k & \textbf{36.36} & 40.72 & \textbf{7.90} \\
\bottomrule
\end{tabular}
\end{table}

Furthermore, the continuous improvement in BoN accuracy underscores the practical value of our designed Verifier in inference-time. Sci-CoE not only enhances the ability of generated candidate solutions but also constructs a reliable internal reward signal, allowing the model to accurately identify correct reasoning trajectories among multiple candidates during inference.

\subsection{Verification Capability}
To evaluate the Verifier independently of the scientific reasoning benchmarks, we consider VerifyBench-Hard (VB-H), SCI-VerifyBench (SCI-VB), and RM-Bench. VB-H focuses on difficult response-verification cases, SCI-VB evaluates verification in scientific domains, and RM-Bench measures the ability to distinguish response quality in reward-model evaluation. We use the official scripts for all three benchmarks and report accuracy of Qwen3-8B backbone.\par
As shown in Table~\ref{tab:verify_results} and Table~\ref{tab:rm_results}, Sci-CoE-Stage2-30k obtains the best score on all three benchmarks, improving VerifyBench-Hard by 0.90 points, SCI-VerifyBench by 1.00 point, and RM-Bench by 0.60 points over the base model, supporting the effectiveness of co-evolution for difficult and science-oriented verification as well as reward-model evaluation.

\subsection{Ablation Study and Analysis}

\noindent \textbf{Impact of Anchored Learning.} We investigated the necessity of the Anchored Learning by skipping Stage 1 and training directly on Stage 2 using unlabeled data. As shown in Table \ref{tab:ablation_result} Index 1, the model significantly lags behind Sci-CoE. In several benchmarks, it even underperforms the baseline. This confirms that although Stage 1 uses a relatively small amount of data, it provides essential training anchors that help the model establish an initial notion of correctness and verification reliability. Notably, in Stage 1, utilizing only 0.4k annotated samples, the model achieves overall performance improvement (Index 2). This confirms that a minimal set of high-quality anchor data is sufficient to successfully bootstrap the fundamental capabilities of both the Solver and Verifier, establishing a solid foundation for subsequent evolution. Furthermore, a comparison between Stage 1-0.4k and Stage 1-4k reveals that while more data yields a better starting point, the system can be successfully bootstrapped with as few as 0.4k samples, demonstrating the framework's robustness.

\noindent \textbf{Effectiveness of Geometric Reward.} We compared the efficacy of the Naive Consensus Reward against our proposed Geometric Reward during Stage 2. As shown in Table \ref{tab:ablation_result} Index 3-4, the model utilizing the Geometric Reward significantly outperforms the Raw Reward version across all benchmarks.\par
To provide a more intuitive explanation, we visualize the verification strategies by projecting their embedding vectors into a 2D space using PCA and representing them in polar coordinates as Figure \ref{fig:cluster} (a-d).
The strategy points of the baseline model are mostly close to red in color, indicating low consensus scores. Meanwhile, they are unevenly distributed and located far from the cluster center, suggesting that the generated strategies are neither reliable nor diverse. Compared to the baseline, the strategy points after Stage 1 exhibit an overall improvement in consensus. However, their angular distribution remains concentrated, indicating limited diversity. For Stage 2 with the Naive Reward, most strategy points are close to green and form several highly dense clusters that cover only a small angular range. This reflects that the model repeatedly generates homogenized and overly simplistic strategies (e.g., simple format checks) to maximize consensus scores, which leads to a significant loss of diversity. In contrast, under the Geometric Reward, strategy points are uniformly distributed along the polar angle while maintaining high reliability (i.e., smaller radial distances). This demonstrates that the geometric reward successfully encourages the Verifier to explore orthogonal verification perspectives, thereby constructing a more robust evaluation system.

\noindent \textbf{Quantitative Dynamics of Reward Components.}
Beyond qualitative visualization, we analyze the quantitative metrics in Figure \ref{fig:cluster}(e-g). The Naive Reward model achieves high consistency ($r^{\text{con}}$) but at the severe cost of diversity ($r^{\text{div}}$). In contrast, the Geometric Reward mechanism achieves a high level of diversity and reliability while also maintaining a decent consistency score. This reveals that our geometric reward acts as a structural regularizer. By penalizing angular redundancy in the latent space, it prevents the model from falling into local optima where the Solver and Verifier prefer simple reasoning trajectories. The balanced improvement across consistency, reliability, and diversity is the key driver behind Sci-CoE's superior generalization on complex scientific questions.

\subsection{Impact of Verification Signals}
To validate the design of our Solver-Verifier co-evolution and address whether external supervision could be simplified, we conduct ablations on the source of the reinforcement learning reward signal: (1) Qwen235B-Direct-Judge, where a massive 235B teacher model directly evaluates solutions without any verification strategy (acting as a direct distillation baseline); and (2) Policy Self-Judge, where the policy model generates both the solution and the verification strategy, but then executes the evaluation itself instead of relying on an external judge. \par
As shown in Table \ref{tab:judge_ablation}, both alternative mechanisms fail to improve the model and instead lead to performance degradation compared to the base model. In complex scientific reasoning, intermediate steps are lengthy and highly specialized. Even a 235B parameter model struggles to reliably verify these solutions directly. Surprisingly, the Policy Self-Judge model slightly outperforms the 235B Direct Judge, despite having only 8B parameters. This compelling result demonstrates that generating an explicit verification strategy is intrinsically valuable. By formalizing the evaluation criteria, a small model equipped with a strategy can construct a more robust judgment than a massive model relying on black-box intuition.

\subsection{Results Across Model Scales}
\noindent \textbf{Cross-scale Applicability.}
Table~\ref{tab:3.5_series_results} shows that Sci-CoE can be used across the Qwen3.5 family from 0.8B to 9B parameters. Qwen3.5-2B exhibits the most consistent improvement: the 18k model exceeds its baseline by 3.03 percentage points on GPQA-Diamond, 0.48 points on MMLU-Pro, and 0.44 points on UGPhysics. Qwen3.5-4B primarily benefits on GPQA-Diamond, improving by 3.03 points after Stage 1 and 2.53 points after Stage 2.\par

\noindent \textbf{Model--Data Difficulty Matching.}
All Qwen3.5 scales are trained on the same data pools, which were originally constructed around the capabilities of Qwen2.5-7B and Qwen3-8B. The results suggest that this fixed distribution is best matched to the intermediate-capacity models: 2B improves consistently across all three benchmarks, and 4B obtains a substantial GPQA-Diamond gain. In contrast, the smaller and less stable gains for 9B suggest that many training questions may be insufficiently challenging for its stronger initial policy, while the Stage~1 degradation of 0.8B indicates that part of the long-form training distribution may exceed its effective reasoning and completion capacity.\par

\noindent \textbf{Evaluation Budget for Small Models.}
All Qwen3.5 models are evaluated with max\_tokens=32,768. This common limit controls evaluation cost but disproportionately affects the 0.8B and 2B models, which more frequently enter repetitive or ineffective reasoning loops and fail to produce a final answer before the limit. Although the officially recommended budget for these small models is 81,920 tokens, adopting it would substantially increase evaluation time. Their absolute scores therefore reflect both reasoning quality and the ability to terminate within the fixed budget.\par

\begin{figure}[t]
\vspace{-6pt}
    \centering
    \includegraphics[width=\columnwidth]{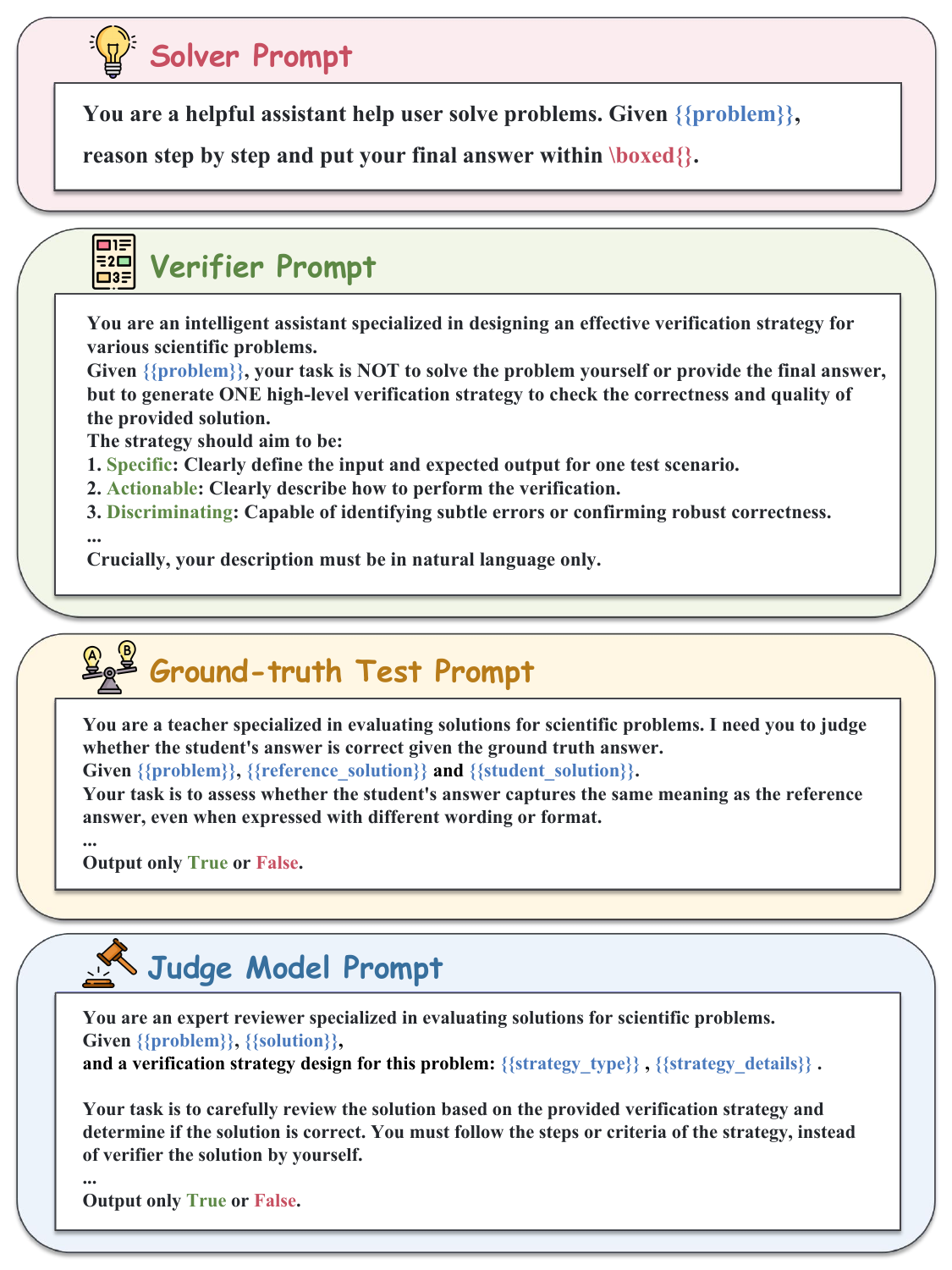}
\vspace{-6pt}
\caption{Prompts for the Solver, Verifier, GT-test, and Judging Model in Sci-CoE.}
\label{fig:prompt}
\end{figure}

\subsection{Case Study}

\begin{figure}[t]
\vspace{-6pt}
    \centering
    \includegraphics[width=\columnwidth]{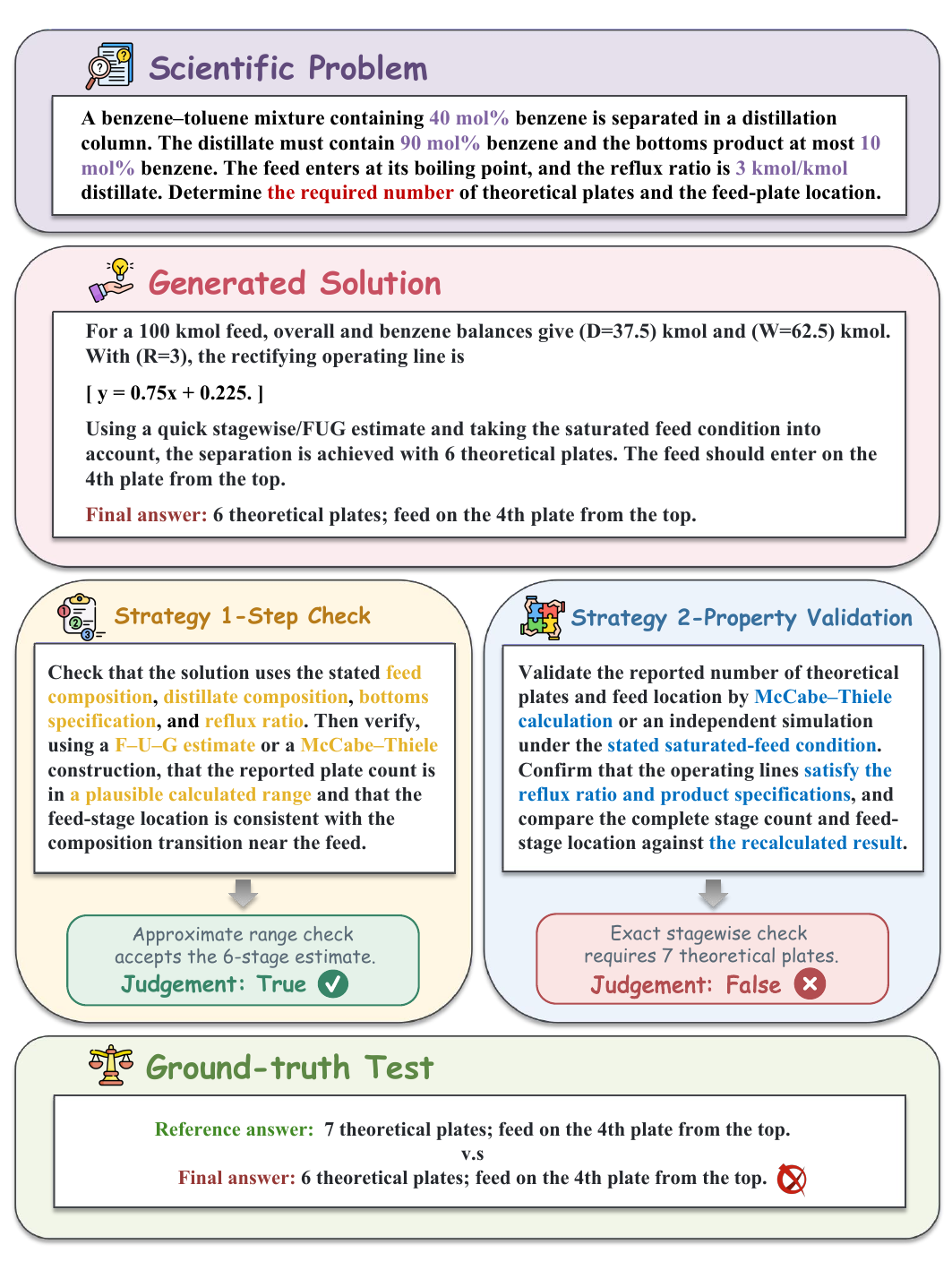}
\vspace{-6pt}
\caption{Case study of complementary strategy verification.}
\label{fig:case}
\end{figure}

Figure~\ref{fig:prompt} shows prompt template of Solver, Verifier, GT-Test, and Judge Model. Figure~\ref{fig:case} shows an example of our solution--strategy evaluation. For the same candidate solution, the broad step-check strategy returns True: the material balances, operating data, and a coarse stage estimate appear plausible. In contrast, the property-validation strategy returns False after an independent stagewise check reveals that the reported six plates are insufficient. The ground-truth test likewise rejects the candidate because the reference requires seven theoretical plates. Thus, the two strategies on a single solution expose both a false positive and a discriminating rejection.\par
This disagreement is informative rather than a failure of the framework. Sci-CoE samples strategies with different semantic emphases, so that a candidate is assessed from multiple scientific perspectives instead of being rewarded by a single, potentially superficial criterion. During co-evolution, solutions that repeatedly satisfy the collection of reliable and diverse strategies obtain a stronger and more stable learning signal; the Solver is consequently driven toward reasoning trajectories that survive all complementary checks, while the Verifier is trained to produce strategies that remain both discriminative and non-redundant. The geometric reward is important in this loop: it discourages the Verifier from collapsing to only permissive strategies such as Strategy 1, allowing the eventual agreement of multiple complementary strategies to represent stronger evidence of solution correctness.
\section{Conclusion}
We present Sci-CoE, a scientific co-evolving framework that improves LLMs’ scientific reasoning under minimal supervision. A key insight is that verification strategies form a structured and learnable space, whose reliability and diversity can be encouraged through geometric modeling. Experimental results demonstrate that Sci-CoE improves reasoning accuracy and robustness, and scales effectively to large unlabeled data. We acknowledge the following limitations. Due to a limited budget, we only trained models with up to eight-billion parameters. Additionally, Sci-CoE currently relies on an external judging model to execute verification strategies, which introduces additional computational cost and potential bias. We believe Sci-CoE represents a meaningful step toward self-evolving scientific reasoning systems and opens new directions for learning reliable reasoning without supervision.
\twocolumn


%\section{}

%\ifCLASSOPTIONcompsoc
  %\section*{Acknowledgments}
%\else
  %\section*{Acknowledgment}
%\fi
\section*{Acknowledgments}
This work was supported by the New Generation Artificial Intelligence-National Science and Technology Major Project under Grant No. 2026ZD0128600, the China Postdoctoral Science Foundation under Grant No. 2025M781483.

\ifCLASSOPTIONcaptionsoff
  \newpage
\fi



%\bibliographystyle{IEEEtran}
\bibliographystyle{IEEEtran}
\bibliography{main}


%\end{IEEEbiography}


%\newpage



%\vfill


\end{document}